# Marathon Environments: Multi-Agent Continuous Control Benchmarks in a Modern Video Game Engine

**Joe Booth**
Vidya Gamer, LLC
joe@joebooth.com

**Jackson Booth**
Summit Sierra High School, Seattle
jboost.si@mysummitps.org

## Abstract

Recent advances in deep reinforcement learning in the paradigm of locomotion using continuous control have raised the interest of game makers for the potential of digital actors using active ragdoll. Currently, the available options to develop these ideas are either researchers' limited codebase or proprietary closed systems. We present Marathon Environments, a suite of open source, continuous control benchmarks implemented on the Unity game engine, using the Unity ML-Agents Toolkit. We demonstrate through these benchmarks that continuous control research is transferable to a commercial game engine. Furthermore, we exhibit the robustness of these environments by reproducing advanced continuous control research, such as learning to walk, run and backflip from motion capture data; learning to navigate complex terrains; and by implementing a video game input control system. We show further robustness by training with alternative algorithms found in OpenAI.Baselines. Finally, we share strategies for significantly reducing the training time.

## 1. Introduction

Video games make substantial use of real time animation. Because video games are user driven, gameplay programmers must work closely with motion capture directors and animators to deal with 'edge cases' between the player and non-player characters and/or the environment. This is an expensive, painful, and never-ending process because of the complexity of the problem space. Furthermore, animation is motion captured, which requires specialist post-processing, which is tedious and time-consuming.

Recent advances in reinforcement learning have provided algorithms that can learn character animation from scratch (Brockman et al., 2016; Todorov, Erez, & Tassa, 2012) and which adapt to a wide variety of environmental challenges (Mnih et al., 2015). Further research has shown that agents can learn from motion capture, DeepMimic (Peng et al., 2018) and from videos, SFV: Reinforcement Learning of Physical Skills from Video (Peng et al., 2018).

If this research can be shown to be transferable into professional video game engines, then this development would have the potential to reduce costs and increase democratization, while improving realism and the range of user interaction. However, commercial video game engines such as Unity3D[1] and Unreal[2] have to take into account many constraints, such as the need to work on many platforms (mobile devices, web, pc/mac, Xbox, PlayStation, and Nintendo), support networking, perform high-end rendering, and maintain an accessible workflow. The physics engines in game engines are optimized for stability and performance at the expense of realism (Erez, Tassa, & Todorov, 2015).

In this paper, we present the core suite of environments and their integration into Unity ML-Agents Toolkit (Juliani et al., 2018). Next, we outline our experiments to reproduce training from motion capture data, to implement a video game controller, and to navigate complex environments. Finally, we describe our integration and benchmarks with OpenAI.Baselines (Dhariwal, et al., 2017) and share our optimization strategies.

## 2. Related work

The MuJoCo physics engine (Todorov et al., 2012) is a popular choice for researchers for building continuous control benchmarks and is used by OpenAI.Gym (Brockman et al., 2016) and the DeepMind Control Suite (Tassa et al., 2018). While MuJoCo is a powerful physics engine, its rendering engine is limited, it is not open source, and, to the best of our knowledge, it has not been implemented in a commercial video game.

The MuJoCo Unity Plugin (Todorov, 2018) enables a MuJoCo physics simulation to be rendered using the Unity Game Engine. It has a number of limitations: the physics simulation needs to run as a separate PC program, it has limited integration with Unity's core object system, and it is proprietary.

---

[1] www.unity3d.com
[2] www.unrealengine.com

Unity ML-Agents Toolkit (Juliani et al., 2018) is an open source framework for machine learning using the Unity platform. It includes some continuous control samples such as a version of ant and humanoid.

Analog RL Locomotion (Booth, 2018) demonstrated that some continuous control environments could be implemented using a commercial game engine; however, it is not open source.

DeepMotion[3] is a proprietary physics platform aimed at game developers wanting to use active ragdolls trained with reinforcement learning. It is the most advanced proposal in this space and supports the leading game engines (Unity & Unreal). However, it is not open source.

Roboschool[4] is a set of open source continuous control benchmarks for OpenAI.Gym (Brockman et al., 2016). Implemented using the open source Bullet physics engine, Roboschool removes the requirement for a Mujuco (Todorov et al., 2012) license. However, it has limited rendering capability and is limited in terms of appeal for game developers.

## 3. Marathon Environments

Marathon Environments[5] consists of four continuous control environments. Each environment parses an xml script containing the agent model definition; this is converted into a Unity object. The xml scripts are slightly modified versions of the xml scripts from the DeepMind Control Suite (Tassa et al., 2018) for Hopper, Walker, Humanoid, and OpenAI.Roboschool for Ant.

All environments are Multi-Agent environments by which we mean they contain multiple instances of a single agent. In each training step the environment collects an observation for each agent instance. For clarity, there is no competition or communication between the agent instances. Unless specified, all environments contain 16 agents.

### 3.1 Integration with Unity ML-Agents Toolkit

Marathon Environments was released alongside Unity ML-Agents Toolkit v0.5[6]. All environments have a single ML-Agent brain, with continuous observations and continuous actions. There are no visual observations. Each action correlates to one motorized joint axis. Multi-axis joints are implemented using a nested joint strategy.

Reward, termination, and observation functions are influenced by DeepMind Control Suite (Tassa et al., 2018) and OpenAI.Roboschool. The one exception was for Humanoid as we implemented a phase function in the reward to enhance training speed.

Each environment is a separate Unity scene with a prefab for each agent type and contains one or more pre-trained Tensorflow models. A custom agent class, which inherits from MarathonAgent.cs, is used to define the behavior of that agent. See Supplementary Materials A for further details.

---

[3] https://www.deepmotion.com
[4] https://github.com/openai/roboschool
[5] https://github.com/Unity-Technologies/marathon-envs
[6] https://blogs.unity3d.com/2018/09/11/ml-agents-toolkit-v0-5-new-resources-for-ai-researchers-available-now/

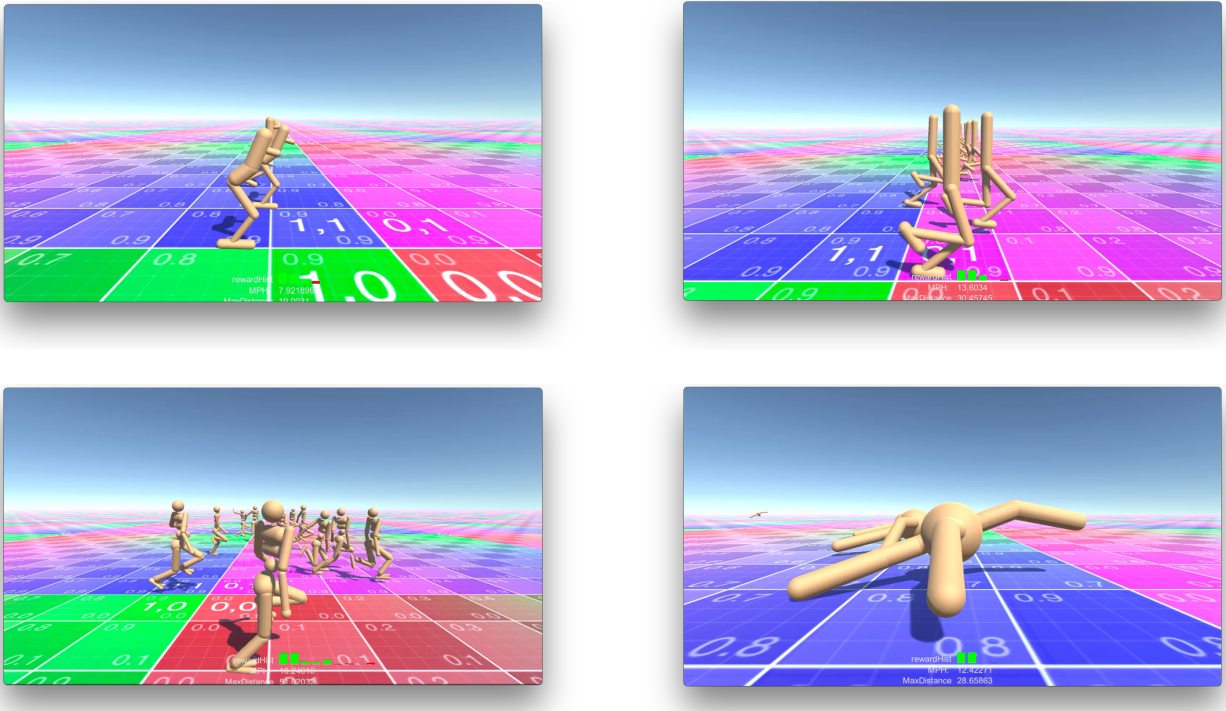

*Figure 1: The four Marathon Environments.*

*Top left: Hopper Environment. Top right: Walker Environment. Bottom left: Humanoid Environment. Bottom right: Ant Environment.*

**The Hopper** agent has 4 joints/actions and 31 observations. The reward function has a positive reward signal for the pelvic velocity and pelvis uprightness. Similarly, there is a negative reward signal for the effort of the current action state and a penalty if the height is below 1.1m. The termination function triggers if a non-foot labeled body part collides with the terrain, if the height falls below <.3m, or if the head tilts more than 0.4 units.

**The Humanoid** agent has 21 joints/actions and 88 observations. The reward function has a positive reward signal for the pelvic velocity and pelvic uprightness, a negative reward signal for the effort of the current action state, and a penalty if the height is below 1.2m. It also adds additional reward based on the phase cycle of its legs. The termination function triggers if a non-foot labeled body part collides with the terrain.

**The Walker** agent has 6 joints/actions and 41 observations. The reward function has a positive reward signal for the pelvic velocity and pelvis uprightness. There is a negative reward signal for the effort of the current action state and a penalty if the height is below 1.1m. The termination function triggers if a non-foot body part collides with the terrain.

**The Ant** agent has 8 joints/actions and 53 observations. The reward function has a positive reward signal for the pelvic velocity, a negative reward signal for the effort of the current action state, and a negative signal if the joints are at their limit. The termination function triggers if the body of the ant rotates more than 0.2 units from upright.

| Environment | Training Steps | Training Time | Steps Per Second | Observations | Actions |
|---|---|---|---|---|---|
| Hopper | 4,800,000 | 36m 31s | 2,190 | 31 | 4 |
| Walker | 4,800,000 | 37m 14s | 2,149 | 43 | 6 |
| Humanoid | 16,000,000 | 3h 37m | 1,229 | 92 | 21 |
| Ant | 4,800,000 | 35m 55s | 2,359 | 54 | 8 |

*Table 1: Training speeds. Compares wall clock training time, and agent training steps per second across environment.*

**Training Performance**

We compared training performance between each environment, training 16 concurrent agents (see Table 1). We found that the steps per second decreased with the increased complexity of the agent.

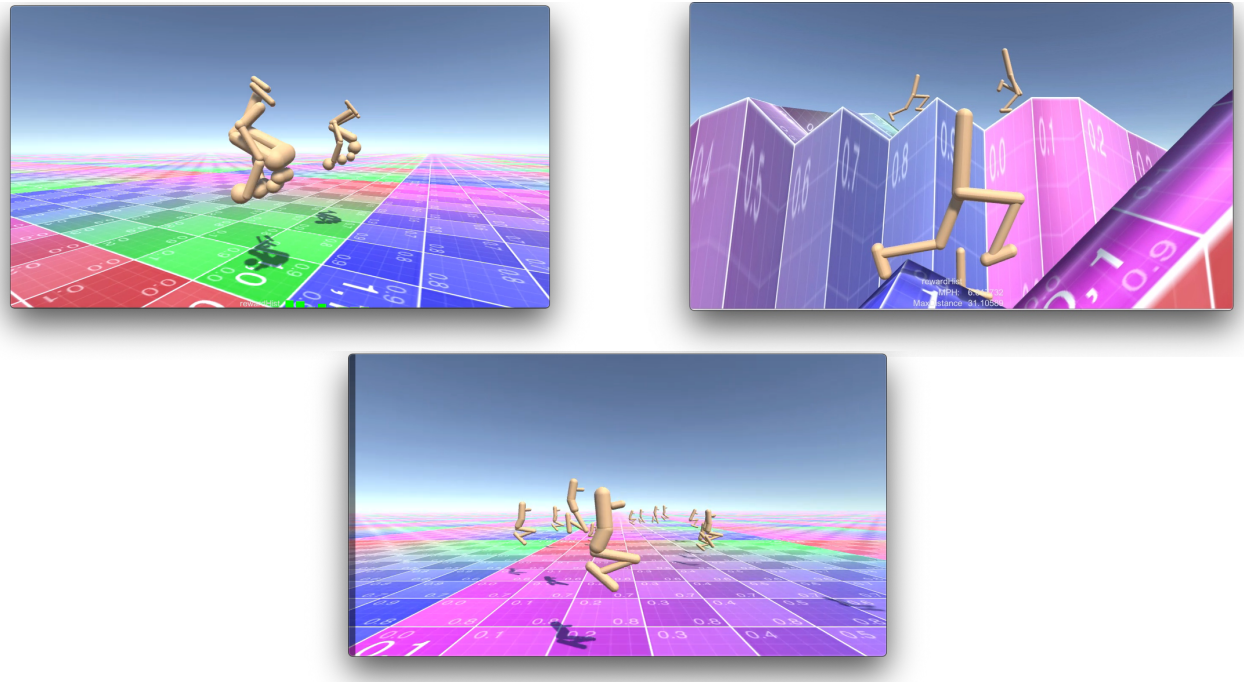

*Figure 2: Additional Experiments. Top left: Active Ragdoll Style Transfer ~ Learning Backflip from motion capture data.*

*Top right: Active Ragdoll Assault Course. Bottom left: Active Ragdoll Controllers.*

## 4. Additional Experiments

As well as reproducing Continuous Control Benchmarks, a number of additional experiments have been produced that use Marathon Environments as a base. The goal of these additional experiments is for the resulting improvements from these experiments to be included in future versions of Marathon Environments.

### 4.1 Active Ragdoll Style Transfer: Learning Locomotion from Motion Capture Data

The goal of these experiments was to confirm that research into training Continuous Control Humanoids from motion capture data will transfer to Unity and Marathon Environments, DeepMimic (Peng, Abbeel, et al., 2018). Source code, animations, and downloadable demos of these experiments can be found on GitHub[7].

**Method:** Experiments were performed with both a standard Unity Humanoid, modified using Unity's built-in Ragdoll tool, and Marathon Man, a modified version of Marathon Environments' Humanoid which was adapted to work with Motion Capture data and has modified joints. We used the standard Unity animation functions to cycle through and record slices of animation. The reward function is based on the distance from the reference animation at each step. To improve training speed, 64 agents were trained in parallel, and a modified version of the Academy script was used so that actions were only applied to the model every 1 in 5 physics steps. The beta version of Unity 2018.3 was also used, as this release of Unity includes improvements to the PhysX engine.

**Results:** While we were able to train the Unity Humanoid, the results were not as high quality as those of DeepMimic (Peng, et al., 2018). On closer inspection, we found that the auto-generated ragdoll had problems with self-collision. Because the Marathon Environments Humanoid had successfully trained to walk without motion capture data, we switched to using Marathon Man. This change allowed us to train against a walking animation with relative ease; however, we found we needed to modify the joints to train running. Additionally, we added independent sensors for the heel and toes. We found that with the modified joints and sensors we were able to train running and improve the performance of walking.

Backflip proved much more challenging. In one experiment, the agent completed half the backflip but without enough height to avoid landing on its head. We addressed this through tuning the termination function to include early termination when the agent was in a state greater than a predefined distance from the reference animation. In another experiment, the agent aborted the backflip and landed on its arms and legs and then returned to a standing position. This fail scenario was due to the agent learning to avoid early termination and maximize the number of steps needed to receive a reward. Despite these challenges we did achieve training backflip when we updated to the Beta version of Unity 2018.3, which includes improvements to the underlying PhsyX physics engine.

---

[7] https://github.com/Sohojoe/ActiveRagdollStyleTransfer

Our experiments of learning locomotion from motion capture data also included interesting findings for rate of learning. It took 50,000,000 training steps for the agent to first complete a backflip, and we continued training to 128,000,000 training steps to improve stability. This compares to the average of 60,000,000 training steps required for DeepMimic (Peng, et al., 2018) and implies that we have room to improve training performance. Our wall clock training speed was 25hrs, implying that we have a 4x training speed improvement over DeepMimic (Peng, et al., 2018) which required 48hrs.

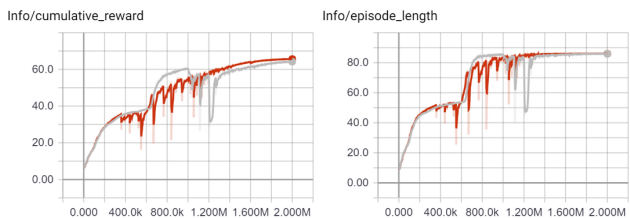

*Figure 3: Training curves for Learning Backflip from Motion Capture Data. Two independent runs. Each training step collects observations for 64 agents for a total training length of 128,000,000 steps.*

### 4.2 Active Ragdoll Controllers: Implementing Player Controllers

The motivation of these experiments was to explore how a typical video game player controller implementation performs when driving a continuous control agent trained through reinforcement learning. The target audience for these experiments is video game designers and developers. Source code, animations, and downloadable demos of these experiments can be found on Github[8].

**Method:** All experiments are 2D and used the Hopper or Walker agents. Experiments included both continuous input (a float value from -1 to 1) and discrete inputs (left, right, no-op, jump, jump+left, jump+right).

For the continuous control experiments, we mapped the Unity game engine controller input strength to the agent's target velocity whereby the human player has control over the agents target direction and velocity. For training, we randomly choses the agent's target velocity between -1 and +1. The target velocity was reselected on a variable time basis between 40 to 240 time steps. We also trained using fixed target velocities of -1, 0 and +1.

For the discrete control experiments we mapped controller input to agent actions of left, right and no-action (stationary). For training, we randomly chose the target action with a probability of 40% choose left, 40% choose right, 20% choose stationary. We also ran experiments adding the additional actions of jump, jump + left, jump + right. During training there was a probability of 25% choose jumping action.

We include the subjective summary by one of our researchers who has 30+ years of video game development experience. We chose this methodology due to the early nature of this work and the prototype feel of the experiments.

> The discrete controller experiments felt satisfying and organic due to the relationship between the input and the movement of the agent. For example, when moving at top speed to the right, on holding the left input, the agent response is contextual depending on its current content. If a jump is in progress, the agent will re-position its weight to stop and reverse on contact with the terrain. If the agent is preparing to jump right, it will abort the jump and perform a left jump. These behaviors are very time consuming to create using traditional game development techniques as they require branch points within animations and controllers. The one area of concern is that the agent would not sit still; it always has some movement, almost like a hyperactive child unable to sit still in their seat. I would like to see further development in environmental challenges, such as variable height terrain and jump challenges.

*Figure 3: Subjective Summary by Experienced Game Developer.*

**Results:** We expected the continuous control experiments to give a higher sense of control due to the continuous mapping of the controller to the target velocity of the agent. However, it was strikingly apparent that the discrete controller agent was more responsive and gave the player more control. We tried training the continuous controller agent using only three inputs (of -1, 0, 1) however, this did not improve the results. This suggests that the lack of control is due to the tuning of the Unity input controller (how physical inputs are mapped to the logical input value). We chose to focus efforts on the discrete controller agent, due to the faster progress and reduced complexity.

The discrete controller agent was more responsive and provided a better sense of control (see Figure 3). The discrete controller agent learned behavior mappings not explicit in the design (see Figure 3), which give the discrete controller agent the higher sense of control we expected from the continuous controller agent. The discrete controller agent also demonstrated contextual behavior, such as adapting its weight to a change input during a jump.

---
[8] https://github.com/Sohojoe/ActiveRagdollControllers

### 4.3 Active Ragdoll Assault Course: Learning to navigate challenging environments.

The goal of these experiments was to validate that adaptability to environmental challenges such as those demonstrated in Emergence of Locomotion Behaviours in Rich Environments (Heess et al., 2017), would transfer to Marathon Environments. Source code, animations, and downloadable demos of these experiments can be found on GitHub[9].

**Method:** All experiments were 2D and used the Hopper or Walker agents. Experiments included manually modifying the terrain, manually adding box objects to the environment, and using an adversarial network to dynamically generate the terrain based on the inverse of the agent's reward signal. We experimented with and without height observations.

**Results:** We showed that our agents were able to train to navigate complex terrains, and that successful strategies in the reference paper transferred to our experiments. We found that the agent was able to learn to navigate basic terrains, even without height observations. We found that the most optimal training was with height observations, and the adversarial network to dynamically generate the terrain.

### 4.4 Marathon Environment Baselines

The original intention of the following experiments was to see if the OpenAI.Baseline algorithms could improve training speed for our Style Transfer experiments. We expected to find a significant performance improvement using algorithms optimized for GPU; however, we found the inverse, in that every algorithm that supported multi-agent learning performed better on CPU training. In addition, we found alternative ways to optimize ML-Agents Toolkit. Our methods and findings may be of interested to researchers looking to use ML-Agents Toolkit and OpenAI.Baselines (Dhariwal, et al., 2017). Source code can be found on Github.[10]

**Method:** First, we created a new repository that contains source code for ML-Agents Toolkit, Marathon Environments, OpenAI.Baselines and Stable-Baselines. To make it easier for others to use this repository, we created pre-built MacOS and Windows environments for Hopper and Walker for solo agent training, 16 agent training, and solo inference mode, all of which can be downloaded from the release page of Github.

In addition, we created Python scripts for invoking solo and multi-agent ML-Agents Toolkit environments from the command line, along with the required modifications to support the execution of these scripts. This includes unity_vec_env.py, which handles mapping the ML-Agents Toolkits' multi-agent implementation to the requirements expected from OpenAI.Baselines (Dhariwal, et al., 2017).

OpenAI.Baselines (Dhariwal, et al., 2017) uses two different strategies for concurrent agent training. The first strategy is concurrently running multi-environments in which MPI is used to handle communication between multiple environments with single agents. To invoke multi-environments, either wrap the command line, Python invoke, with `mpiexec python ....`, or use the `--num_env=XX` command line option. The second strategy is a multi-agent environment, wherein a single environment contains multiple agents which train concurrently.

All our experiments used Marathon Environment Hopper over 1m training steps. All multi-agent experiments were performed using the same Unity ML-Agent environment. We used the MuJoCo hyper-parameters; the only modification made was to divide the default `nsteps` by the number of concurrent agents, which in this experiment was 16.

For collecting the ML-Agent.PPO data, we modified the Marathon Environments hyper-parameters to limit to 1m training steps and to match Baseline's hyper-parameters as closely as possible (see supplementary materials).

Unless specified, all experiments were performed on a 6 core i7 8700k @ 3.70 GHz, with a NVIDIA GTX 1080 GPU, using tensorflow-gpu v1.7.1, optimized for AVX2 instructions[11].

For comparisons between MPI multi-environment and multi-agent environments, we used four concurrent MPI environments and the 16-agent environment. We tested these on both the PC architecture, as shown above, and a MacBook Pro 2016.

**Limitations**: We experienced the following limitations when using OpenAI.Baselines with Unity ML-Agents Toolkit:
1. Save/Load is broken when using normalized training, meaning it is not possible to load trained models.
2. MPI is broken on Windows. We were able to train standard OpenAI.Gym environments with MPI, but found errors when training Unity ML-Agents Toolkit environments.
3. Multi-agent training is only supported by a subset of OpenAI.Baseline algorithms (PPO2 CPU & GPU, A2C CPU & GPU, ACKTR CPU only).
4. A2C and ACKTR did not report score as the Baselines.Monitor wrapper does not support vectorized environments.

**Preliminary Results:**

**MPI Multi-Environment vs Multi-Agent:** On PC we observed a 7.5 x performance improvement in wall clock time (87% per agent), while on MacOS we observed a 6x speed improvement (50% per agent). While we were limited in our capacity to test with MPI, our results indicate that Multi-Agent has the most potential for improving training time, especially when considering the findings reported in the performance section below.

---

[9] https://github.com/Sohojoe/ActiveRagdollAssaultCourse
[10] https://github.com/Sohojoe/MarathonEnvsBaselines

[11]https://github.com/fo40225/tensorflow-windows-wheel/tree/master/1.7.1/py36/GPU/cuda92cudnn71avx2

**CPU vs GPU training speed:** We found training with CPU significantly faster than training with GPU (ranging from 30% to 3x improvement as shown in Table 2). This was unexpected; we had hoped that OpenAI.Baseline would outperform CPU training with a GPU. We don't believe that this is caused by the Unity game engine competing for GPU resources, as we re-ran experiments with rendering disabled, PhysX forced to CPU, and found this had little impact on the training time. By measuring the wall clock time the python algorithm spends between collecting samples and batch training, we found that the increase was during the batch training phase. The collecting samples phase took an increase of 60% (700ms for GPU, 430ms for CPU), whereas the batch training phase took an increase of 500% (2,030ms for GPU, 330ms for CPU). More research is needed to confirm whether the root cause is due to an inefficiency in Baselines implementation of vectorized environments, or a quirk related to our specific GPU, the version of Tensorflow we used, or an edge case specific to environment size and hyperparameters.

| Algorithm | CPU Training Speed | GPU Training Speed | GPU Delta |
|---|---|---|---|
| OpenAI.Baselines.A2C | 4m 51s (291s) | 7m 38s (458s) | 158% of CPU |
| OpenAI.Baselines.PPO2 | 7m 55s (475s) | 23m 43s (1423s) | 300% of CPU |
| Unity ML-Agents.PPO | 6m 58s (418s) | 9m 23 (563s) | 135% of CPU |

Table 2: CPU training speed outperforms GPU. The algorithm column states the framework and algorithm. The CPU and GPU training speed columns state the training time with and without a GPU. The GPU Delta column states the percentage of time the GPU took to train vs training with the CPU.

**ML-Agents Toolkit vs OpenAI.Baselines:** Our preliminary findings demonstrated that OpanAI.Baselines algorithms have the potential to outperform Unity ML-Agents Toolkit. Baselines.PPO2 scored 50% higher, while training 14% slower. A2C & ACKTR trained 33% faster.

| Algorithm | CPU Training Speed | Score | Speed vs ML-Agents | Score vs ML-Agents |
|---|---|---|---|---|
| Unity ML-Agents.PPO | 6m 58s (418s) | 455 | - | - |
| OpenAI.Baselines.A2C | 4m 51s (291s) | n/a | 65% | n/a |
| OpenAI.Baselines.ACKTR | 4m 41s (281s) | n/a | 67% | n/a |
| OpenAI.Baselines.PPO2 | 7m 55s (475s) | 700 | 114% | 154% |

Table 3: ML-Agents Toolkit vs OpenAI.Baselines training performance. The algorithm column states the framework and algorithm. The score column states the final scores. A2C and ACKTR did not report scores due to limitations of their implementation. The Speed vs ML-Agents column details the relative training time as a percentage of ML-Agents Toolkit's training time. The Score vs ML-Agents column details the relative training score as a percentage of ML-Agents Toolkit's score.

### 4.5 Optimization Strategies

The following optimizations strategies were implemented in our Style Transfer experiments. Our goal was to explore strategies to reduce training wall clock time.

**Increasing the number of concurrent Agents:** Stooke & Abbeel, 2018, demonstrated that the PPO algorithm has the potential to run up to 512 concurrent environments with small variations in the results on discrete Atari environments. By default, Marathon Environments uses 16 concurrent agents. We found that we were able to scale to 64 concurrent agents and improve wall clock training time by 45%. The only modification made was to scale `max_steps`. Unity ML-Agents Toolkits' total training steps is the product of the number of concurrent agents and the `max_steps` brain setting in the yaml config file.

**Headless Mode**: Unity ML-Agents Toolkit supports headless mode, sometimes called Server Mode, by specifying the `--no-graphics` command line option.

**Modified Academy to Support Physics Only Steps:** Marathon Agents increases the number of physics steps per second to 200-500, depending on environment. One reduces the frequency of learning steps by setting the `Decision-Frequency` parameter in the ML-Agent agent's settings; however, the action step is triggered with every physics step, impacting performance. By modifying the Academy script and adding a `PhysicsOnlySteps,` we reduced the action frequency to that of the training frequency. The combined impact of this optimization and the `--no-graphics` optimization improved wall clock training time by 50%.

**Optimized Tensorflow for CPU:** As discussed in the OpenAI.Baselines sections, installing an optimized version of Tenesorflow and running the algorithms in CPU mode significantly improved training wall clock time.

## 5. Challenges

### 5.1 PhysX.
We faced three major challenges with the PhysX physics engine:

**Bugs and Compromises** made by the PhysX engine, especially with regard to joints, was our primary challenge. For example, the Configurable Joint object combines settings between the Y and Z axis, limiting the ability of the joint to map to all requirements. A second problem is the general stability of joints and the need to experiment with different joint strategies and settings. The general advice of the community is to use Configurable Joints or nested joints instead of Character Joints.

**Poor Documentation** of the PhysX physics engine was the second challenge we faced. PhysX is developed by NVIDIA and implemented into Unity by Unity engineers. There is rudimentary documentation in the Unity manual and SDK documentation. However, for more technical matters, one is left to search through support forums or explore the native PhysX documentation. To make matters worse, because PhysX has been around for a number of years, many posts and documentation sources are out of date.

**Our Lack of Experience** with PhysX and physics engines was our third area of challenge. Working with Quaternions is not natural, having to convert between left-handed and right-handed coordinate systems is confusing, and the multiple coordinate systems within Unity cause extra complications (world, object, joint, local, global).

### 5.2 Overfitting the Reward Function.

A second area of challenge is the tendency of reinforcement learning algorithms to overfit. We found that development was slowed down because bugs would not always be apparent as the agent learned to compensate and solve a buggy implementation of the environment. This would not become apparent until scaling to a more complex environment where the agent would fail to learn leaving us unsure whether the problem was with the newer environment or with something lower in the stack.

These overfitting challenges were compounded by the PhysX problems.

## 6. Conclusion

In this paper, we presented Marathon Environments, a suite of open source continuous control benchmarks implemented in a modern commercial video game engine. We demonstrated the robustness of these environments and the ability to transfer cutting-edge reinforcement learning research into a video game setting. We discussed various strategies to reduce training time, and we documented how these environments can be used with OpenAI.Baselines (Dhariwal, et al., 2017). Collectively, these experiments show that reinforcement learning research is transferable to a video game engine, and that future investment towards digital actors using active ragdoll is justified.

## Acknowledgements

We would like to thank Arthur Juliani, Jeffrey Shih, and Leon Chen for their contributions to and testing of the integration into ML-Agents Toolkit, and Dr. Nicole Swedberg and Anna Booth for their contributions to, and feedback in writing this paper.

# Supplementary Materials

## A. Detail on using the AgentReset, StepReward, and TerminateFunction functions when modifying agents.

`AgentReset()` is used to initialize the agent. It should set the callbacks for `StepRewardFunction`, `TerminateFunction`, and `ObservationsFunction`. The developer should add model elements to the `BodyParts` list and call `SetupBodyParts()` to initialize the body parts. This enables callbacks to leverage helper functions; for example, `GetForwardBonus(pelvis)` calculates a bonus based on the body part's distance from the forward vector.

`StepReward()` returns a `float` with the reward for the current action step. Helper functions include the following: `GetVelocity(pelvis)` returns the velocity of the specified body part; `GetEffort()` returns the sum of the current actions (one can pass a list of body parts to ignore); and `GetJointsAtLimitPenality()` returns a penalty for actions which are at their limit.

`TerminateFunction()` returns `true` if a termination condition is met. Termination functions help improve training speed by reducing the agent's exposure to unhelpful observations. Helper termination functions include `TerminateNever()`, which never terminates (always returns `false`) and `TerminateOnNonFootHitTerrain()`, which returns `true` if a body part that is not a foot collides with the terrain. Body parts are defined in the function `OnTerrainCollision()`. Some agents required the lower leg body parts to be labeled as `foot` as they protrude through the foot geometry and create false positive terminations.

## B. Training Hyperparameters for the Training Performance in section 3.

```
DeepMindHumanoidBrain:
    normalize: true
    num_epoch: 3
    beta: 0.01
    time_horizon: 1000
    batch_size: 2048
    buffer_size: 20480
    gamma: 0.995
    max_steps: 2e6
    summary_freq: 1000
    num_layers: 2
    hidden_units: 512
DeepMindHopperBrain:
    beta: 1.0e-2
    epsilon: 0.20
    gamma: 0.99
    lambd: 0.95
    learning_rate: 1.0e-3
    num_epoch: 3
    time_horizon: 128
    summary_freq: 1000
    use_recurrent: false
    normalize: true
    num_layers: 2
    hidden_units: 90
    batch_size: 2048
    buffer_size: 10240
    max_steps: 3e5
    use_curiosity: true
    curiosity_strength: 0.01
    curiosity_enc_size: 256
DeepMindWalkerBrain:
    beta: 1.0e-2
    epsilon: 0.20
    gamma: 0.99
    lambd: 0.95
    learning_rate: 1.0e-3
    num_epoch: 3
    time_horizon: 128
    summary_freq: 1000
    use_recurrent: false
    normalize: true
    num_layers: 3
    hidden_units: 41
    batch_size: 2048
    buffer_size: 10240
    max_steps: 3e5
    use_curiosity: true
    curiosity_strength: 0.01
    curiosity_enc_size: 256
OpenAIAntBrain:
    beta: 5.0e-3
    epsilon: 0.20
    gamma: 0.99
    lambd: 0.95
    learning_rate: 1.0e-3
    num_epoch: 3
    time_horizon: 128
    summary_freq: 1000
    use_recurrent: false
    normalize: true
    batch_size: 2048
    buffer_size: 10240
    num_layers: 3
    hidden_units: 53
    max_steps: 3e5
```

## C. Training Hyperparameters for Marathon Environment Baselines.

```
DeepMindHopperBrain:
    learning_rate: 3.0e-3
    num_epoch: 10
    time_horizon: 128
    summary_freq: 1000
    use_recurrent: false
    normalize: true
    num_layers: 2
    hidden_units: 64
    batch_size: 2048
    buffer_size: 10240
    max_steps: 62500
```